%% file: main.tex
\documentclass[runningheads]{llncs}

\usepackage{eccvabbrv}
\usepackage{graphicx}
\usepackage{booktabs}
\usepackage{url}
\usepackage{enumitem}
\usepackage{wrapfig}
\usepackage{multirow}
\usepackage{comment}
\usepackage{makecell}
\usepackage{amsmath,bm}
\usepackage{float}
\usepackage{marvosym}

\usepackage[accsupp]{axessibility}

\usepackage{hyperref}
\usepackage{orcidlink}

\begin{document}

\def\logo{\makebox[22pt][l]{\raisebox{-0.9ex}{\includegraphics[height=20pt]{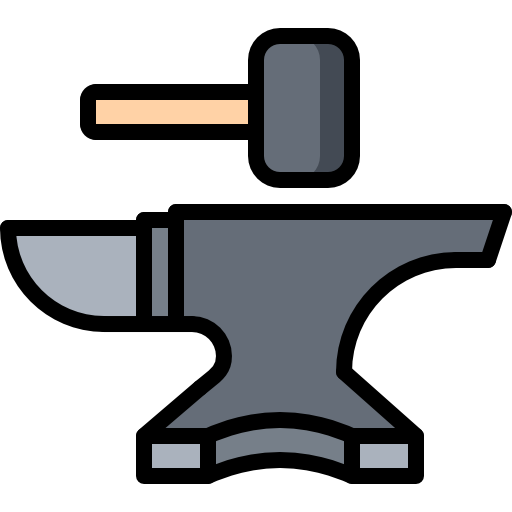}}\hspace{20pt}}}
\definecolor{color1}{RGB}{143,71,109}
\definecolor{color2}{RGB}{196,123,145}
\definecolor{color3}{RGB}{227,179,189}
\definecolor{color4}{RGB}{255,222,173}
\newcommand{\methodname}{Forge4D}

\title{\logo \textcolor{color1}{For}\textcolor{color2}{ge}\textcolor{color3}{4D}: Feed-\textcolor{color1}{For}ward \textcolor{color3}{4D} Human Reconstruction and Interpolation from Uncalibrated Sparse-View Videos}

\titlerunning{Forge4D}

\author{Yingdong Hu$^{\ast}$\inst{1}\orcidlink{0009-0002-8708-1001}
\and
Yisheng He$^{\ast}$\textsuperscript{\Letter}\inst{2}\orcidlink{0000-0002-4029-5494} \and
Jinnan Chen\inst{3}\orcidlink{0009-0007-9964-2132} \and
Weihao Yuan\inst{2}\orcidlink{0000-0002-1362-3747} \and
Kejie Qiu\inst{2}\orcidlink{0000-0002-2516-3020}  \and
Zehong Lin\inst{4}\orcidlink{0000-0002-9503-2464}  \and
Siyu Zhu\inst{5}\orcidlink{0000-0003-0293-0044}  \and
Zilong Dong\inst{2}\orcidlink{ 0000-0002-6833-9102}  \and
Steven Hoi\inst{2}\orcidlink{0000-0002-4584-3453}\and
Jun Zhang\inst{1}\orcidlink{0000-0002-5222-1898}}

\authorrunning{Y.~Hu et al.}

\institute{The Hong Kong University of Science and Technology, Hong Kong SAR, China\and
Tongyi Lab, Alibaba Group, Hangzhou, China
\and
National University of Singapore, Singapore
\and
School of Data Science, Lingnan University, Hong Kong SAR, China\and
Fudan University, Shanghai, China\\
\email{yingdong.hu@connect.ust.hk, ethanheysh@gmail.com}
}

\maketitle
\footnotetext[1]{$^{\ast}$ indicates equal contribution.}
\footnotetext[2]{\textsuperscript{\Letter} Corresponding author. Contact:\texttt{ethanheysh@gmail.com}}

\input{sec/0_abs}
\input{sec/1_intro}
\input{sec/2_related_works}
\input{sec/3_method}
\input{sec/4_experiment}

\input{sec/5_conclusion}

\bibliographystyle{splncs04}
\bibliography{main}
\appendix
\input{sec/x_appendix}
\end{document}

%% file: sec/0_abs.tex
\begin{abstract}
Instant reconstruction of dynamic humans and 3D human-object interactions from uncalibrated sparse-view videos is
critical for numerous downstream applications. Existing methods, however, are either limited by the slow reconstruction speeds or incapable of generating novel-time representations. To address these challenges, we propose \textit{\methodname}, a feed-forward 4D human-object reconstruction and interpolation model that efficiently reconstructs temporally aligned representations from uncalibrated sparse-view videos, enabling both novel-view and novel-time synthesis. Our model simplifies the 4D reconstruction and interpolation problem as a joint task of streaming 3D Gaussian reconstruction and dense motion prediction. 
For the task of streaming 3D Gaussian reconstruction, we introduce learnable state tokens to enforce temporal consistency in a memory-friendly manner.
For novel-time synthesis, we design a novel motion prediction module to predict dense motions for each 3D Gaussian between two adjacent frames.
To overcome the lack of the ground truth for dense motion supervision, we formulate dense motion prediction as a dense point matching task and introduce a self-supervised \textit{retargeting loss} to optimize this module. An additional occlusion-aware \textit{optical flow loss} is introduced to ensure motion consistency with plausible human and object movement, providing stronger regularization. Extensive experiments demonstrate the effectiveness of our model on both in-domain and out-of-domain datasets. Code and models will be made publicly available. 
\keywords{Feed-forward 4D Human Reconstruction \and 4D Gaussian Splatting \and Novel View Synthesis}
\end{abstract}

%% file: sec/1_intro.tex
\section{Introduction}
Instant 4D human-object reconstruction from uncalibrated sparse-view video streams is essential for various application scenarios, including real-time livestreaming~\cite{livestreaming}, sports broadcasting, augmented/virtual reality (AR/VR)~\cite{ar}, articulation modeling~\cite{3dpt,autoconnect,rig,phys}, and immersive holographic communication~\cite{telealoha,li2025panolam,LAM_Sig25}. However, this task remains challenging due to the inherent difficulty of simultaneously recovering accurate human-object geometry and dense motion trajectories from unposed sparse-view video streams, while maintaining the real-time interactivity required for practical applications. For example, holographic communication systems demand high interactability, while sports broadcasting requires the ability to present novel views at any time for enhanced viewing experiences and precise evaluation of athletic performance. 

Existing works~\cite{mega,stg2024,dualgs} typically rely on iterative optimization over entire dense-view video sequences for each scene. These approaches depend heavily on calibrated camera parameters and suffer from prolonged training durations required for 4D representation convergence. 
Meanwhile, recent visual geometry grounded models~\cite{vggt,dust3r,pi3} have enabled intermediate 3D point cloud reconstruction and camera pose estimation from arbitrary long uncalibrated image sequences in a feed-forward manner. However, the inherent limitations of point cloud representations restrict their ability to synthesis photorealistic novel-view images.
Subsequent works~\cite{anysplat,noposplat} have extended feed-forward reconstruction models to predict static 3D Gaussians (3DGS), enabling photorealistic novel-view synthesis. Nevertheless, these methods remain incapable of handling dynamic scenes and synthesizing novel-time images.

\begin{figure*}[t]
    \centering  \includegraphics[width=.91\linewidth]{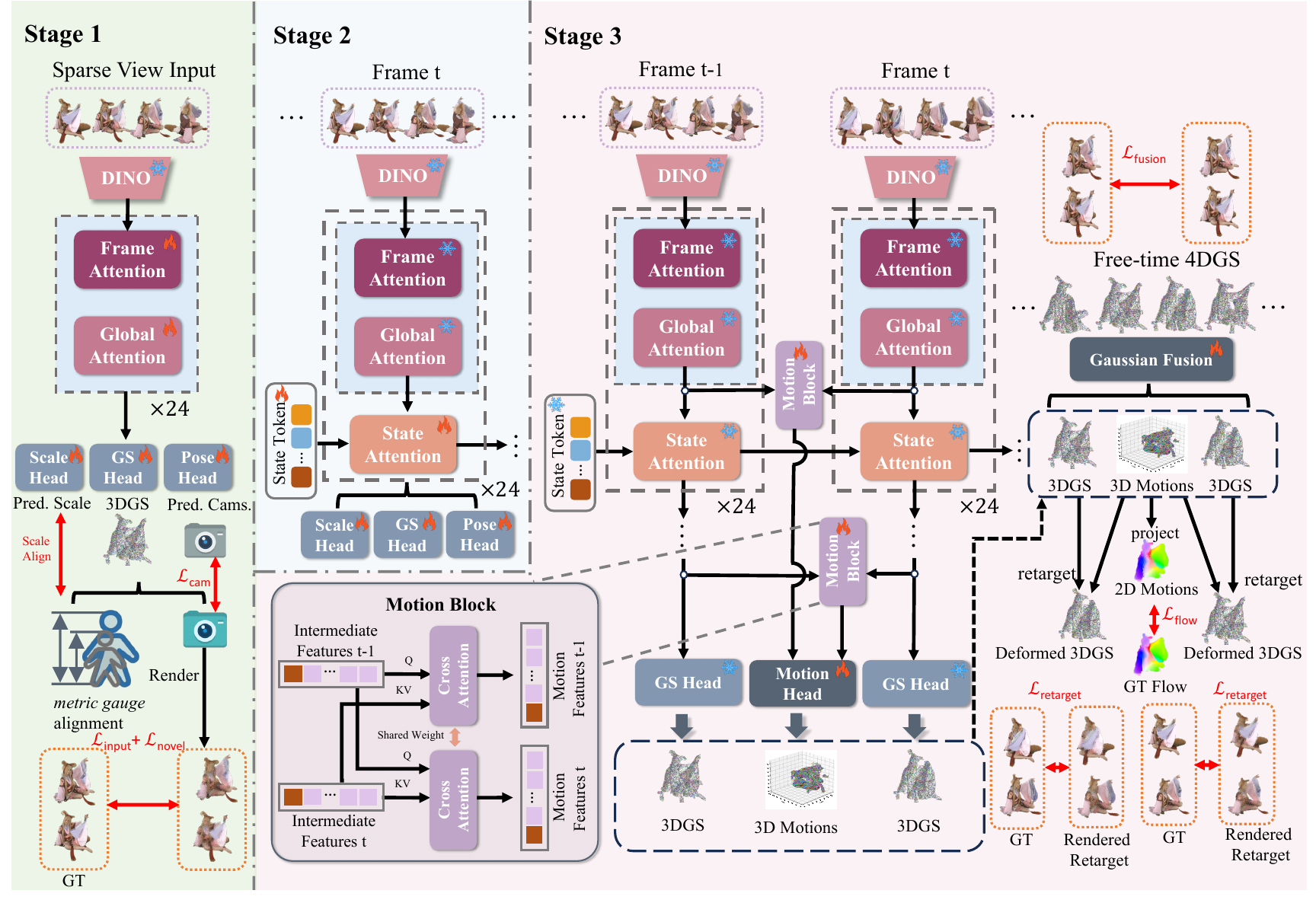}
    \caption{The overall pipeline of \textit{\methodname}.
    It is trained in three stages: (1) static feed-forward 3D Gaussian reconstruction stage; (2) a streaming stage temporally aligned via state tokens; and (3) a feed-forward 4D reconstruction stage that predicts dense motion for each 3D Gaussian and interpolates free-time 3DGS using an occlusion-aware fusion process.
    }
    \label{fig:pipeline}
\end{figure*}
In this work, we propose \methodname, the first \textit{feed-forward} model for \textit{metric-scale} 4D human-object reconstruction that enables \textit{novel-view} and \textit{novel-time} synthesis from \textit{uncalibrated sparse-view} videos in an \textit{efficient streaming manner}. Our framework enables: 1) efficient reconstruction of temporally consistent 3D Gaussian assets in real-world metric scale from streaming sparse-view video inputs, and 2) accurate frame-wise dense 3D motion prediction for human-object subjects and 4D interpolation for novel-time synthesis. 
To achieve these goals, we decompose the 4D reconstruction and interpolation problem into two tasks: streaming 3D Gaussian reconstruction and dense motion prediction. This design offers two advantages: 1) it simplifies the problem for feed-forward regression, and 2) the reconstructed streaming 3DGS provide visual supervision for accurate dense motion prediction. 
Specifically, for streaming 3D Gaussian reconstruction, we leverage the pretrained knowledge prior from the large 3D reconstruction model VGGT~\cite{vggt} and adapt it to predict streaming key-frame 3D Gaussian assets. This adaptation is non-trivial due to two major challenges. First, a scale discrepancy exists between VGGT's output and the real-world metric scale inherent in ground-truth camera extrinsics, causing fundamental misalignment and unstable optimization with novel-view photometric loss. Second, naively feeding VGGT with multiple video frames suffers from long reconstruction duration, low interactivity, and out-of-memory (OOM) issues due to increasing image tokens for global attention. To address the scale issue, we propose to maintain a \textit{metric gauge} and force the model to generate a temporally consistent scale in Sec.~\ref{stage1}, which not only improves the stability under novel-view supervision, but also enables metric measurement. For the efficiency and OOM problem, we decompose the spatial and temporal dimensions of sparse-view videos and propose state tokens in Sec.~\ref{stage2} to iteratively incorporate temporal information in a streaming manner. 

For dense motion prediction and novel-time synthesis, we propose a dense motion prediction module in Sec.~\ref{stage3} to facilitate 3D representation synthesis at arbitrary intermediate timestamps. In contrast to prior approaches that merely leverage middle-frame photometric supervision, our key insight is to formulate the task of dense motion prediction as a 3D Gaussian point-matching problem.
However, there is no ground truth dense 3D motion for supervision of this module. Therefore, we propose a novel \textit{retargeting loss} that projects current 3DGS to adjacent frames with the predicted dense motion and supervises the rendered images against ground truth. This regularization optimizes the dense motion in a self-supervised manner. For a stronger regularization, we also propose an \textit{occlusion-aware optical flow loss}
to enhance the plausibility of predicted motions by supervise them with 2D optical flows from a prior model. Given the dense motion between two timestamps, we deform the dynamic 3DGS from the two nearest frames under a constant velocity assumption for novel-time synthesis. These deformed representations are then merged using an occlusion-aware fusion MLP to reduce 3DGS numbers and avoid flickering artifacts. Experimental results on benchmark datasets demonstrate the efficiency and effectiveness of the proposed framework. 

The main contributions of this work are summarized as:
\begin{itemize}[leftmargin=*]
\item We propose the first feed-forward model for metric-scale 4D human-object reconstruction from uncalibrated sparse-view videos, enabling novel-view synthesis and novel-time 4D interpolation in an efficient streaming manner.
\item Our model simplifies this task by decomposing it into sequential streaming 3D Gaussian prediction and dense motion estimation. The novel \textit{metric gauge} regularization, \textit{retargeting loss}, and \textit{occlusion-aware optical flow loss} stabilize the optimization and significantly improve motion prediction, photorealistic novel-view and novel-time synthesis.
\item We introduce a novel motion-guided, occlusion-aware Gaussian fusion method for 3D Gaussian interpolation, enabling novel-time synthesis and effectively mitigating flickering and jittering artifacts caused by temporal redundancy in dynamic 3D Gaussian representations.
\end{itemize}

%% file: sec/2_related_works.tex
\section{Related Work}

\noindent\textbf{Dynamic Scene Reconstruction and Streaming.}
Dynamic scene reconstruction from multi-view videos is crucial for numerous real-world applications. Prior methods primarily focus on optimizing a unified 4D representation to match dense multi-view 2D observations either by incorporating temporal dimensions into spatial coordinates~\cite{diffuman,mega,4dgs,4drotorgs,ex4dgs} or by deforming 3D representations from keyframes using dynamic factors~\cite{dynamic3dgs,gsflow,dualgs,freetimegs,splatter}. Another line of research assumes causal inputs and reconstructs per-frame 3D representations in a streaming manner~\cite{queen,dass,instantgsstream,3dgstream}. However, these methods suffer from a limited reconstruction speed and are sensitive to the number of input views. In contrast to these iterative optimization-based approaches, \textit{\methodname} introduces an efficient feed-forward model that reconstructs the entire 4D scene in a single forward pass from uncalibrated sparse videos, significantly enhancing interactivity and applicability to downstream tasks.

\noindent\textbf{Feed-Forward Reconstruction.}
Recent advances in visual geometry models have demonstrated the capability of deep neural networks for 3D reconstruction from multi-view images in a feed-forward manner. DUSt3R~\cite{dust3r}, VGGT~\cite{vggt}, and $\pi^3$~\cite{pi3} enable direct regression of camera poses and 3D point maps in the first frame's coordinate space. To enable photorealistic novel-view synthesis, another line of works~\cite{pixelsplat,mvsplat,gpsgaussian,evagaussian,generalizable,gbcsplat,monofusion} directly predict static 3DGS ~\cite{freesplat,gamba,mvgamba,depthsplat,flare,mvgaussian} or textured meshes~\cite{pshuman} from calibrated multi-view images. To alleviate the reliance on camera calibration, NoPosplat~\cite{noposplat} and AnySplat~\cite{anysplat} propose to reconstruct 3DGS from uncalibrated multi-view images. However, all these methods are limited to per-timestamp static reconstruction and cannot synthesize novel-time 3D representations. Although L4GM~\cite{l4gm} extends the static Gaussian reconstruction framework~\cite{LGM} to feed-forward 4D reconstruction, it suffers from low-resolution reconstructions and poor generalization on real-world subjects. Concurrent to our work, recent methods~\cite{4dgt,dgslrm,movies} target 4D Gaussian reconstruction from monocular calibrated videos, and StreamSplat~\cite{streamsplat} further extends to uncalibrated ones. However, these methods neither explore the critical connection between 4D reconstruction and 3D point matching nor formulate motion learning as an explicit geometric correspondence problem, resulting in suboptimal performance. In contrast, our model is specifically designed to recover detailed geometry, appearance, and dense frame-wise motion for human performance from uncalibrated multi-view videos. By reformulating 4D reconstruction as a 3D Gaussian point matching task and introducing specialized losses for motion retargeting and occlusion-aware flow alignment, our approach achieves superior reconstruction quality and temporal consistency.

\noindent\textbf{Template-Based Human Reconstruction and Animation.}
Several methods~\cite{lhm,LAM_Sig25,panolam,guava,gaussianavatar,gaussianavatars,anigs,hfhuman} reconstruct animatable avatars from single-view or sparse observations with parametric human priors like SMPL-X~\cite{smplx} and FLAME~\cite{FLAME:SiggraphAsia2017}, which are then driven by estimated SMPL-X parameters from videos during inference. However, they fail to model the dynamic garments and human-object interaction and results in stiff and unnatural rendered videos due to the fundamental limitation of parametric human model. The drifting errors of SMPL-X parameters estimation from videos also hinder the stable performance. In contrast, \textit{\methodname} regresses pixel-aligned 3DGS representations with dense motion prediction to capture dynamic subjects, completely eliminating reliance on human templates while maintaining compatibility with dynamic clothing and complex human-object interactions.

%% file: sec/3_method.tex
\section{Method}
\subsection{Overview}

This work utilizes a transformer-based model $\mathcal{D}_{\text{4D}}$ for feed-forward reconstruction and novel-time interpolation of dynamic 4D Gaussian $\mathcal{G}_{4D}$ from $n$ sparse uncalibrated videos $\{\bm{I}^t_i\}_{i=0,t=0}^{n,k}$ with a consistent video length of $k$, which can be expressed in the form of:

\begin{equation}
    \mathcal{G}_{4D} = \mathcal{D}_{\text{4D}}(\{\bm{I}^t_i\}_{i=0,t=0}^{n,k}).
\end{equation}
However, the strong entanglement between object geometry and motion makes the direct regression of 4D Gaussians challenging. To address this issue, we propose to decompose the problem into a streaming 3D Gaussian reconstruction task and a dense motion prediction task, and develop a progressive training pipeline. As shown in Fig.~\ref{fig:pipeline}, the proposed pipeline is composed of three stages: 1) a feed-forward static 3D reconstruction stage to reconstruct static 3DGS in the real-world metric scale, 2) a streaming dynamic reconstruction stage to reconstruct streaming 4D Gaussians in an efficient and memory-friendly way, and 3) a dense motion prediction and Gaussian fusion stage to enable novel-time synthesis. In the first stage, a novel metric gauge calculation method is proposed to align the backbone output scale with the real-world scale, which is critical to more stable supervision of novel views. While directly applying the 3D Gaussian reconstruction pipeline to each time stamp suffers from scale misalignment between different times, the main purpose of the second stage is to align different scales across different time stamps. To this end, we propose a state token to encode information from former frames and interact with the immediate frame efficiently, while being memory-friendly.
In the final stage, a novel motion prediction module is proposed, together with a novel dual frame \textit{retargeting loss} and \textit{occlusion-aware optical flow loss} that marry the task of dynamic Gaussian motion prediction to the task of point matching. To enable novel-time synthesis, an additional occlusion-aware Gaussian fusion procedure is proposed for better dynamic 3D Gaussian interpolation.

\subsection{Stage 1: Feed-forward 3D Gaussian Reconstruction}
\label{stage1}
Recent advances in large 3D reconstruction models have demonstrated remarkable capabilities in recovering colored point maps and camera poses from a set of uncalibrated images. However, the point cloud representation limits the capability for photorealistic synthesis. To address this issue, we propose a feed-forward 3D Gaussian reconstruction model $\mathcal{D}_{\text{3D}}$ that leverages the geometry prior within these foundations, while introducing a 3D Gaussian prediction branch for photorealistic rendering. Specifically, we use a pre-trained VGGT as our backbone and predict pixel-aligned 3DGS with an additional DPT~\cite{dpt} head as $\mathcal{G}^t_{3D} = \mathcal{D}_{\text{3D}}(\{\bm{I}^t_i\}_{i=0}^{n})$, where $\mathcal{G}^t_{3D}=\{\bm{P}_i^t, \bm{O}_i^t, \bm{C}_i^t, \bm{Q}_i^t, \bm{S}_i^t\}_{i=0}^n$, with $\bm{P}_i^t \in \mathbb{R}^{3\times H\times W}$, $\bm{O}_i^t \in \mathbb{R}_i^{1\times H\times W}$, $\bm{C}^t \in \mathbb{R}^{3\times H\times W}$, $\bm{Q}_i^t \in \mathbb{R}^{4\times H\times W}$, and $\bm{S}_i^t \in \mathbb{R}^{3\times H\times W}$ representing the Gaussian position, opacity, color, rotation, and scale maps, respectively, from view $i$ of size $H\times W$ at time $t$. 

\begin{wrapfigure}[14]{r}{0.43\textwidth}
\centering
  \includegraphics[width=0.45\columnwidth, trim={0cm 0cm 0cm 0cm}, clip]{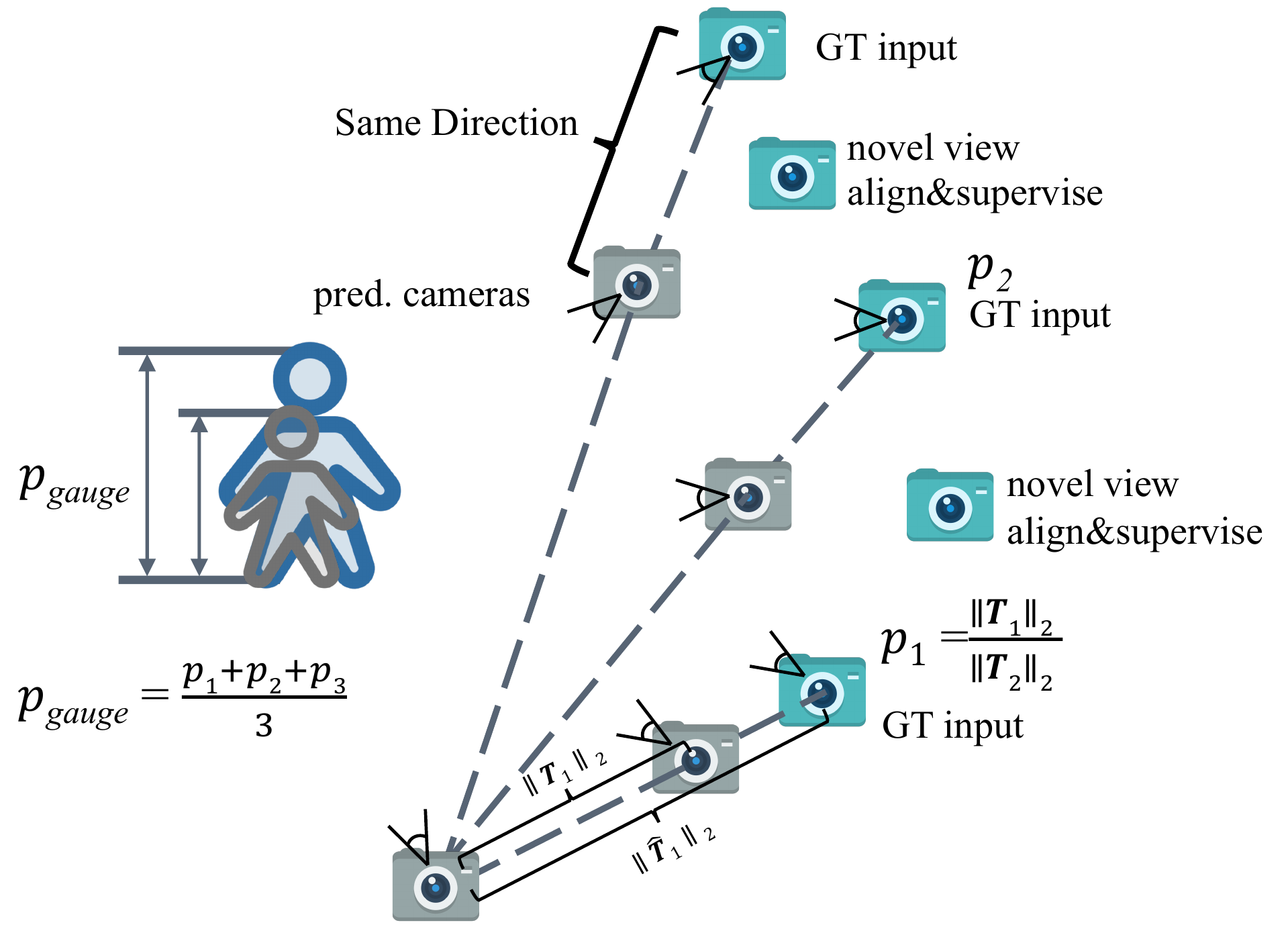}
\caption{Gauge illustration.}
\label{fig:gauge_illustration}
\end{wrapfigure}

To supervise this branch, photometric losses (e.g., L2, SSIM, LPIPS) are applied between the rendered and ground-truth (GT) images. However, a fundamental scale ambiguity occurs between the output scale of VGGT (i.e., normalized point clouds) and the real-world metric scale. Directly applying photometric supervision without addressing this scale discrepancy results in an unstable optimization trajectory and fails to converge to a coherent 3D structure, as further evaluated in Sec. \ref{ablation}. 
To resolve this issue, we introduce a \textit{metric gauge} regularization term  $p_{\text{gauge}}$ to align the scale of the GT novel-view camera extrinsics with the model's internal coordinate system, thereby stabilizing training. This approach is grounded in the key insight that if the model's predicted camera poses and intrinsics are accurate, their difference from the GT poses should be primarily a consistent translation scaling factor, as visualized in Fig.~\ref{fig:gauge_illustration}. Formally, for $n$ input cameras, we calculate the ratio $p_i = {\|\bm{T}_i\|_2}/{\|\Hat{\bm{T}}_i\|_2}$ of translation magnitudes between each predicted camera and its GT counterpart. The novel-view cameras are then scaled using the mean of these ratios, $p_{\text{gauge}} = \frac{1}{n}\sum_{i=1}^{n} p_i$. This factor is also utilized for scale head supervision, which predicts $\hat{p}_{\text{gauge}}$ for metric scale recovery during evaluation. To ensure that the metric gauge accurately represents the scale difference and to simultaneously refine the predicted camera parameters, we propose a comprehensive camera loss:
\begin{equation}
\begin{aligned}
    \mathcal{L}_{\text{cam}} = &\sum_{i=0}^{n}\|\bm{q}_i - \Hat{\bm{q}}_i\|_2 + \sum_{i=1}^{n}\left\| \frac{\bm{T}_i}{\|\bm{T}_i\|_2} - \frac{\Hat{\bm{T}}_i}{\|\Hat{\bm{T}}_i\|_2} \right\|_2 \\& + \sum_{i=1}^{n}|p_i - p_{\text{gauge}}|+|\hat{p}_{\text{gauge}} - p_{\text{gauge}}|,
    \label{eq:camera_loss}
\end{aligned}
\end{equation}
which supervises the model for accurate rotation $\bm{q}_i$, translation direction, consistent relative scaling. and the scale prediction header. Our full training objective then combines this metric-aware camera loss with multi-view photometric supervision. The photometric loss $\mathcal{L}_{\text{input}} = \sum_{i=0}^{n} ( \|\bm{I}_i - \Hat{\bm{I}}_i\|_2 + \lambda_{\text{SSIM}}\text{SSIM}(\bm{I}_i, \Hat{\bm{I}}_i) + \lambda_{\text{LPIPS}}\text{LPIPS}(\bm{I}_i, \Hat{\bm{I}}_i) )$ is applied to the $n$ input views, and a corresponding loss $\mathcal{L}_{\text{novel}}$ is applied to $m$ held-out novel views, ensuring high-fidelity reconstruction across all perspectives.
Thus, the total loss for our feed-forward 3D Gaussian reconstruction model is
$\mathcal{L}_{\text{3D}} = \mathcal{L}_{\text{cam}} + \mathcal{L}_{\text{input}} + \mathcal{L}_{\text{novel}}.$
Supervised by this combined objective, our model not only infers coherent 3DGS from uncalibrated RGB inputs, but also enables metric measuring, which we discuss thoroughly in Supplementary C.

\subsection{Stage 2: Aligning Dynamic Human Streaming with State-token}
\label{stage2}
In stage one, we obtain a feed-forward network for static 3D Gaussian reconstruction from sparse-view images. However, to apply for video scenarios, the output scale of this network is not aligned across timestamps, leading to temporal inconsistency. To address this, one vanilla way is to stack all sparse video tokens for global attention in VGGT, which results in OOM and low reconstruction speed issues. Instead, we decompose the spatial and temporal dimensions of sparse-view videos and introduce a state token to enforce the temporal consistency in an efficient and memory-friendly way. Specifically, we employ a learnable \textit{state token} that iteratively encodes information from all previous frames and broadcasts it to the current frame. These tokens inject temporal information by serving as the Key and Value in a cross-attention layer applied to the current frame's features. Conversely, the token itself is updated by attending to the current frame's features, where it serves as the Query. To further enhance the temporal stability, we extend the metric gauge regularization in Sec.~\ref{stage1} to a temporal form. The global scale factor is now computed over all $n$ cameras and $k$ timestamps as $p_{\text{gauge}} = \frac{1}{(n-1)(k-1)}\sum_{t=1}^{k}\sum_{i=1}^{n} p^t_i$, which is also utilized to supervise the general gauge $\hat{p}^t_{\text{gauge}}$ prediction in the scale prediction header. Consequently, we generalize the camera loss to supervise the temporal cross-attention layers across the entire sequence as:
\begin{equation}
\begin{aligned}
    \mathcal{L}_{\text{cam}} = &\sum_{t=0}^{k}\sum_{i=0}^{n}\|\bm{q}^t_i - \Hat{\bm{q}}^t_i\|_2 + \sum_{t=0}^{k}\sum_{i=1}^{n}\left\| \frac{\bm{T}^t_i}{\|\bm{T}^t_i\|_2} - \frac{\Hat{\bm{T}}^t_i}{\|\Hat{\bm{T}}^t_i\|_2} \right\|_2 \\&+ \sum_{t=0}^{k}\sum_{i=1}^{n}|p^t_i - p_{\text{gauge}}|+ \sum_{t=0}^{k}|\hat{p}^t_{\text{gauge}} - p_{\text{gauge}}|.
    \label{eq:temporal_camera_loss}
\end{aligned}
\end{equation}
This ensures consistent camera rotation, translation direction, and global scale across time.

\subsection{Stage 3: Novel-timestamp 3D Gaussian Synthesis with Dense Motion Prediction and Dynamic Gaussian Fusion}
\label{stage3}
A general 4D representation should enable the synthesis of 2D images from arbitrary camera viewpoints at \textit{any moment} in time. This requires the capability to interpolate the representation to intermediate timestamps beyond the input frames.
In this work, we adopt dynamic 3DGS as our fundamental 4D representation and assume a linear motion model that propagates 3DGS between consecutive frames, following previous work on 4D reconstruction~\cite{freetimegs}. Mathematically, we formulate our 4D Gaussian representation as:
\begin{equation}
\mathcal{G}_{4D} = \{\{\mathcal{G}_{3D}^t\}_{t=0}^{k}, \{\bm{M}_{i,\{1,2\}}^t\}_{i=0, t=0}^{n, k}, \bm{F}_{\theta}\},
\end{equation}
where $\mathcal{G}_{3D}^t$ denotes the 3D Gaussian attribute maps at time $t$, $\bm{M}_{i,\{1,2\}}^t \in \mathbb{R}^{2\times3\times H \times W}$ represents the associated 3D motion field for view $i$, and $\bm{F}_{\theta}$ is a learnable fusion function that adaptively combines Gaussian attributes at novel timestamps in account of occlusion relationships.

To predict the 3D motion map $\bm{M}_{i,\{1,2\}}^t$, we introduce a dense motion prediction block that operates on the static streaming reconstruction described in Sec.~\ref{stage1}. This block predicts a dense, pixel-aligned dual motion field for each 3D Gaussian. Specifically, when processing a new frame at time $t$, the block infers: (1) a backward 3D motion $\bm{M}_{i,1}^t\in \mathbb{R}^{3\times H \times W}$ that warps the current 3DGS (time $t$) to the previous timestamp $t-1$, and (2) a forward 3D motion $\bm{M}_{i,2}^{t-1}\in \mathbb{R}^{3\times H \times W}$ that warps the 3DGS from the previous frame (time $t-1$) to the current frame $t$.
This process is repeated symmetrically when the subsequent frames arrive.
The proposed block comprises the same number of attention blocks as the backbone model. Each motion attention block takes as input the corresponding intermediate features from frames $t$ and $t-1$ produced by the backbone. The output features from all motion attention blocks are aggregated and passed to a motion DPT head to produce the final dual motion prediction $\bm{M}_{i,\{1,2\}}^t$.

Given 3D Gaussian assets at two successive frames $t$ and $t-1$, along with their corresponding motions, the 3DGS at the middle timestamp $t'$ can be acquired by warping these two frames with a constant velocity assumption. This assumption is a first-order simplification of the polynomial 4D Gaussians \cite{stg2024}, and is empirically found to perform adequately for small temporal intervals in our cases. Specifically, for each 3D Gaussian in frame $t$, it is deformed to the middle timestamp by adding a displacement proportional to its temporal distance to time $t$. The 3DGS in frame $t-1$ are also deformed to this middle timestamp $t'$ in the same way. This deforming process can also be represented as:
\begin{equation}
\begin{aligned}
    \bm{P}_i^{t\to t'} &= \bm{P}_i^{t} + |t'-t|\cdot\bm{M}_{i,1}^{t}, \bm{P}_i^{t-1\to t'} \\&= \bm{P}_i^{t-1} + |t'-(t-1)|\cdot\bm{M}_{i,2}^{t-1}.
\end{aligned}
\end{equation}
However, there are no ground truth dense motions for supervision. To effectively train this block and the deformation process, we introduce a novel \textit{retargeting loss} that optimizes the predicted 3D motion using 2D photometric constraints in a self-supervised manner. Specifically, for 3DGS at time $t$, we deform their positions to time $t-1$ with the prediction dense motion as $\bm{P}_i^{t \to t-1} = \bm{P}_i^{t} + \bm{M}_{i,1}^{t}$, while retaining other attributes. This forms an intermediate 3D Gaussian representation $\mathcal{G}_{3D}^{t\to t-1} = \{\bm{P}_i^{t \to t-1}, \bm{O}_i^t, \bm{C}_i^t, \bm{Q}_i^t, \bm{S}_i^t\}_{i=0}^n$. Since $\bm{M}_{i,1}^{t}$ aims to recover the true motion between frames, $\mathcal{G}_{3D}^{t\to t-1}$ should closely align with the ground-truth Gaussians $\mathcal{G}_{3D}^{t-1}$. which can be supervised with rendering consistency.
The retargeting loss is defined as:
\begin{equation}
\begin{aligned}
\mathcal{L}_{\text{retarget}} =& \sum_{i=0}^{n} ( ||\hat{\bm{I}}_i^{t\to t-1} - \hat{\bm{I}}_i^{t-1}||_2 + \lambda_{\text{S}}\text{\small SSIM}(\hat{\bm{I}}_i^{t\to t-1}, \hat{\bm{I}}_i^{t-1}) \\& + \lambda_{\text{L}}\text{\small LPIPS}(\hat{\bm{I}}_i^{t\to t-1}, \hat{\bm{I}}_i^{t-1}) ),
\end{aligned}
\end{equation}
where $\hat{\bm{I}}_i^{t\to t-1} = \mathcal{R}(\mathcal{G}_{3D}^{t\to t-1}, \bm{E}_i, \bm{K}_i)$ and $\hat{\bm{I}}_i^{t-1} = \mathcal{R}(\mathcal{G}_{3D}^{t-1}, \bm{E}_i, \bm{K}_i)$ denote rendered images for view $i$, with $\mathcal{R}$ representing the rendering function for 3D Gaussian Splatting, and $\bm{E}_i$, $\bm{K}_i$ denoting camera extrinsics and intrinsics, respectively.

To ensure real-world plausibility and resolve ambiguities, we incorporate an \textit{occlusion-aware optical flow loss} for a stronger regularization. We compute pseudo-ground-truth flow $\bm{\mu}_i^{t\to t-1}$ using SEA-RAFT~\cite{searaft} and project the predicted 3D motion $\bm{M}_{i,1}^{t}$ to 2D scene flow as $\hat{\bm{\mu}}_i^{t\to t-1}$.
A cyclic consistency mask $\bm{1}_{\text{cyc}}$ penalizes inconsistencies between forward and backward flows and removes occluded regions. The flow loss is defined as:
\begin{equation}
\mathcal{L}_{\text{flow}} = \sum_{i=0}^{n} \bm{1}_{\text{cyc}}(\bm{\mu}^{t\to t-1}_i, \bm{\mu}^{t-1\to t}_i) \cdot ||\bm{\mu}^{t\to t-1}_i - \hat{\bm{\mu}}^{t\to t-1}_i||_2,
\end{equation}
where $\bm{1}_{\text{cyc}}(\bm{\mu}^{t\to t-1}_i, \bm{\mu}^{t-1\to t}_i) = \exp(-\bm{r}^t_i\cdot||\bm{\mu}^{t\to t-1}_i + \bm{\mu}^{t-1\to t}_i[\bm{p}_i^{t} + \bm{\mu}^{t-1\to t}_i]||_2)$ acts as an occlusion-aware weighting term, $\bm{r}^t_i$ is a hyperparameter related to the length of each flow, and $\bm{\mu}^{t-1\to t}_i[\bm{p}_i^{t} + \bm{\mu}^{t-1\to t}_i]$ represents a pixel-wise indexing process. The \textit{retargeting loss} and \textit{occlusion-aware optical flow loss} are combined together as a matching supervision: $\mathcal{L}_{\text{matching}} = \mathcal{L}_{\text{flow}}+\mathcal{L}_{\text{retarget}}$.

\begin{wrapfigure}[13]{r}{0.4\textwidth}
\centering
  \includegraphics[width=0.5\columnwidth, trim={0cm 0cm 0cm 0cm}, clip]{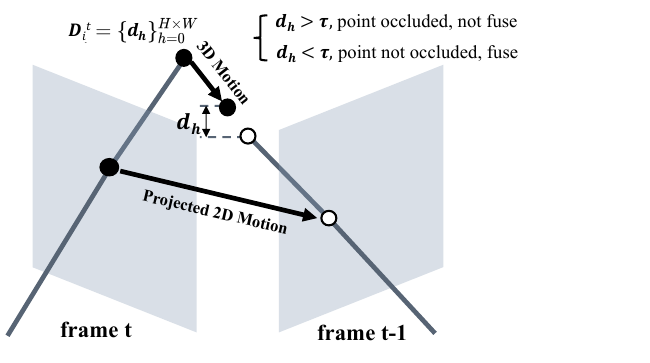}
\caption{Illustration of dual consistency factor. We identify the occluded points with a predicted 2D/3D motion consistency.}
\label{fig:dualconsistencyfactor}
\end{wrapfigure}

Additionally, to naturally fuse the two sets of deformed 3DGS $\mathcal{G}_{3D}^{t\to t'}$, $\mathcal{G}_{3D}^{t-1 \to t'}$ while preserving 3DGS from occluded regions, we further deform the 3D Gaussian with a \textit{dual consistency factor} $\bm{D}_i^{t}$ (or $\bm{D}_i^{t-1}$ for frame $t-1$). This factor is calculated by measuring the distance of the deformed concurrent 3D Gaussian point to the retrieved 3D Gaussian point at the retargeted frame via the projected 2D flow, as shown in Fig.~\ref{fig:dualconsistencyfactor}. This factor serves as the guidance of areas masked in the next frame. 3DGS with a factor larger than a threshold $\tau$ will be kept as occluded 3DGS $\{\bar{\mathcal{G}}_{3D}^{t}, \bar{\mathcal{G}}_{3D}^{t-1}\}$, while the other 3DGS $\{\hat{\mathcal{G}}_{3D}^{t}, \hat{\mathcal{G}}_{3D}^{t-1}\}$ from two nearby timestamps will be merged into one by a two-layer MLP $\bm{F}_{\theta}$ to eliminate temporal redundancy. The final 3D Gaussian assets $\mathcal{G}_{3D}^{t'}$ are merged with the remaining occluded 3DGS and the fused 3DGS as:
\begin{equation}
\mathcal{G}_{3D}^{t'} = \{ \bar{\mathcal{G}}_{3D}^{t\to t'}, \bar{\mathcal{G}}_{3D}^{t-1\to t'}, \bm{F}_{\theta}( \hat{\mathcal{G}}_{3D}^{t\to t'}, \hat{\mathcal{G}}_{3D}^{t-1\to t'})\}.
\end{equation}
This fusion process is supervised by the photometric loss at novel time $t'$, defined as $\mathcal{L}_{\text{fusion}}= \sum_{i=0}^{n} ( \|\bm{I}^{t'}_i - \Hat{\bm{I}}^{t'}_i\|_2 +\lambda_{\text{S}}\text{SSIM}(\bm{I}^{t'}_i, \Hat{\bm{I}}^{t'}_i) + \lambda_{\text{L}}\text{LPIPS}(\bm{I}^{t'}_i, \Hat{\bm{I}}^{t'}_i) )$.
We supervise Stage 3 with the loss function $\mathcal{L}_{4D}=\mathcal{L}_{\text{matching}}+\mathcal{L}_{\text{fusion}}$. 

In this way, our model achieves the task of generalized 4D human-object reconstruction by decomposing it into static 3D Gaussian streaming and dense motion prediction. High-quality novel-view images at any novel time can be acquired by interpolating the predicted dynamic 3DGS and then rendering onto the corresponding image planes.

%% file: sec/4_experiment.tex
\begin{table}[t]
\caption{Quantitative results of novel-view synthesis of static reconstruction. }
\scriptsize
\centering
\resizebox{.9\linewidth}{!}{
\begin{tabular}{cccccccccc}
\hline
\textbf{Dataset} & \textbf{} & \textbf{} && \multicolumn{3}{c|}{\textbf{DNA-Rendering}} & \multicolumn{3}{c}{\textbf{Genebody}} \\
Method           & \multicolumn{2}{c}{Type}    & w. Cam. pose & PSNR$\uparrow$  & SSIM$\uparrow$  & \multicolumn{1}{c|}{LPIPS$\downarrow$} & PSNR$\uparrow$       & SSIM$\uparrow $      & LPIPS$\downarrow$      \\ \hline 
DualGS        & Optimization & 4D*& Yes &18.7408& 0.8932& \multicolumn{1}{c|}{0.1515}      & 16.1721 & 0.7663 &0.1891   \\
D-3DGS      & Optimization   & 4D& Yes &20.7767&0.8944& \multicolumn{1}{c|}{0.1162}      &15.7067 & 0.7981& 0.1790   \\
Queen     & Streaming & 3D& Yes &15.4966&0.8941& \multicolumn{1}{c|}{0.1213}      & 16.4514&0.9484&0.0710\\\hline
GPS-Gaussian     & Feed-Forward & 3D&Yes& 24.2963  &0.9247 & \multicolumn{1}{c|}{0.0867}      & 25.1734 &  0.9346   &  0.0756  \\
NoPoSplat              & Feed-Forward & 3D&\textbf{No}& 11.7632 &0.8092 & \multicolumn{1}{c|}{0.2846}      &13.9554& 0.8721& 0.1664\\
AnySplat              & Feed-Forward & 3D&\textbf{No}& 26.1157 &0.9430 & \multicolumn{1}{c|}{0.1513}      &25.8010&0.9287 &0.1355  \\ \hline
Ours             & Feed-Forward & 4D&\textbf{No}& \textbf{29.8167} &\textbf{0.9606} & \multicolumn{1}{c|}{\textbf{0.0542}}      &  \textbf{28.0819}  &  \textbf{0.9523}  & \textbf{0.0548} \\ \hline
\end{tabular}}
\label{3Dexp}
\end{table}

\begin{table*}[t]
\caption{Quantitative results of novel-time and novel-view synthesis.}
\scriptsize
\centering
\resizebox{.9\linewidth}{!}{
\begin{tabular}{ccccccccc}
\hline
\textbf{Dataset} & \textbf{} && \multicolumn{3}{c|}{\textbf{DNA-Rendering}} & \multicolumn{3}{c}{\textbf{Genebody}} \\
Method           & Type    & w. Cam. pose & PSNR$\uparrow$  & SSIM$\uparrow$  & \multicolumn{1}{c|}{LPIPS$\downarrow$} & PSNR$\uparrow$       & SSIM$\uparrow $      & LPIPS$\downarrow$      \\ \hline 
D-3DGS      & Optimization  & Yes & 20.9158& 0.8994& \multicolumn{1}{c|}{0.1163}      &15.5823& 0.8002& 0.1502\\
SpaceTimeGS           & Optimization   & Yes &17.2189& 0.8879 & \multicolumn{1}{c|}{0.1247} &14.6119&0.8488&0.1805            \\
L4GM     & Feed-Forward &Yes&18.0325  &0.9152& \multicolumn{1}{c|}{0.1367}      & 14.8572& 0.9144& 0.1727 \\ \hline
Ours             & Feed-Forward &\textbf{No}&\textbf{29.0378} &\textbf{0.9566} & \multicolumn{1}{c|}{\textbf{0.0535}}      &\textbf{27.4247} &\textbf{0.9459} & \textbf{0.0601}\\ \hline
\end{tabular}}
\label{4Dexp}
\end{table*}

\section{Experiment}
\label{exp}

\subsection{Experimental Settings}
\textbf{Datasets.} \textit{\methodname} is trained on the DNA-Rendering~\cite{dna} training set, which comprises 2,078 human-object interaction video sequences that exhibit diverse subject ages, appearances, and motion patterns.
4D synthesis is evaluated on three benchmarks: 1) an in-domain held-out test set that contains all sequences from 10 distinct identities in DNA-Rendering, 2) the out-of-domain complete Genebody~\cite{genebody} dataset, and 3) an in-the-wild benchmark composed of two casually captured datasets: SelfCap~\cite{freetimegs} and VVC~\cite{vvc}. For motion and metric prediction, due to a lack of ground truth annotations in real-world datasets, we construct a synthetic dataset, MetaHuman4D, with ground truth annotations for evaluation. The details of MetaHuman4D are in the Supplementary B.

\noindent\textbf{Evaluation Metrics.} The synthesized image quality is measured using standard metrics: PSNR, SSIM, and LPIPS at a resolution of $518 \times 518$, unless specific ones are required by architectural constraints, such as $512 \times 512$ for GPS-Gaussian and L4GM. All models are trained and evaluated using 4 input views with a camera angle of around $45^{\circ}$, which ensures sufficient coverage of the frontal human appearance. The dense motion prediction task is benchmarked using the L2 distance and the retargeted point distance. The metric scale prediction is evaluated by computing the L2 distance between predicted 3D points and their nearest corresponding points on the GT scale mesh.

\noindent\textbf{Baselines.} We establish comparisons under two experimental settings. For novel-view synthesis at input timestamps, we compare against optimization-based methods (DualGS~\cite{dualgs}, Queen~\cite{queen}, D-3DGS~\cite{deformable3dgs}) and feed-forward models (GPS-Gaussian~\cite{gpsgaussian}), along with pose-free feed-forward methods (NoPosplat~\cite{noposplat}, AnySplat~\cite{anysplat}). For novel-time interpolation quality, we compare exclusively with methods capable of generating 3D representations at non-input timestamps: SpaceTimeGS~\cite{stg2024}, D-3DGS~\cite{deformable3dgs}, and L4DM~\cite{l4gm}. We provide comparisons with template-based methods in the Supplementary, noting that the methodological differences prevent a direct comparison.

\noindent\textbf{Implementation Details.} We initialize the backbone using pre-trained weights from VGGT. The scale head, 3D Gaussian position offset head, color offset head, scene attention blocks, and motion blocks are zero-initialized. See more details in the Supplementary C.

\subsection{Evaluation}
\begin{figure}[t]
    \centering
    \includegraphics[width=0.9\linewidth]{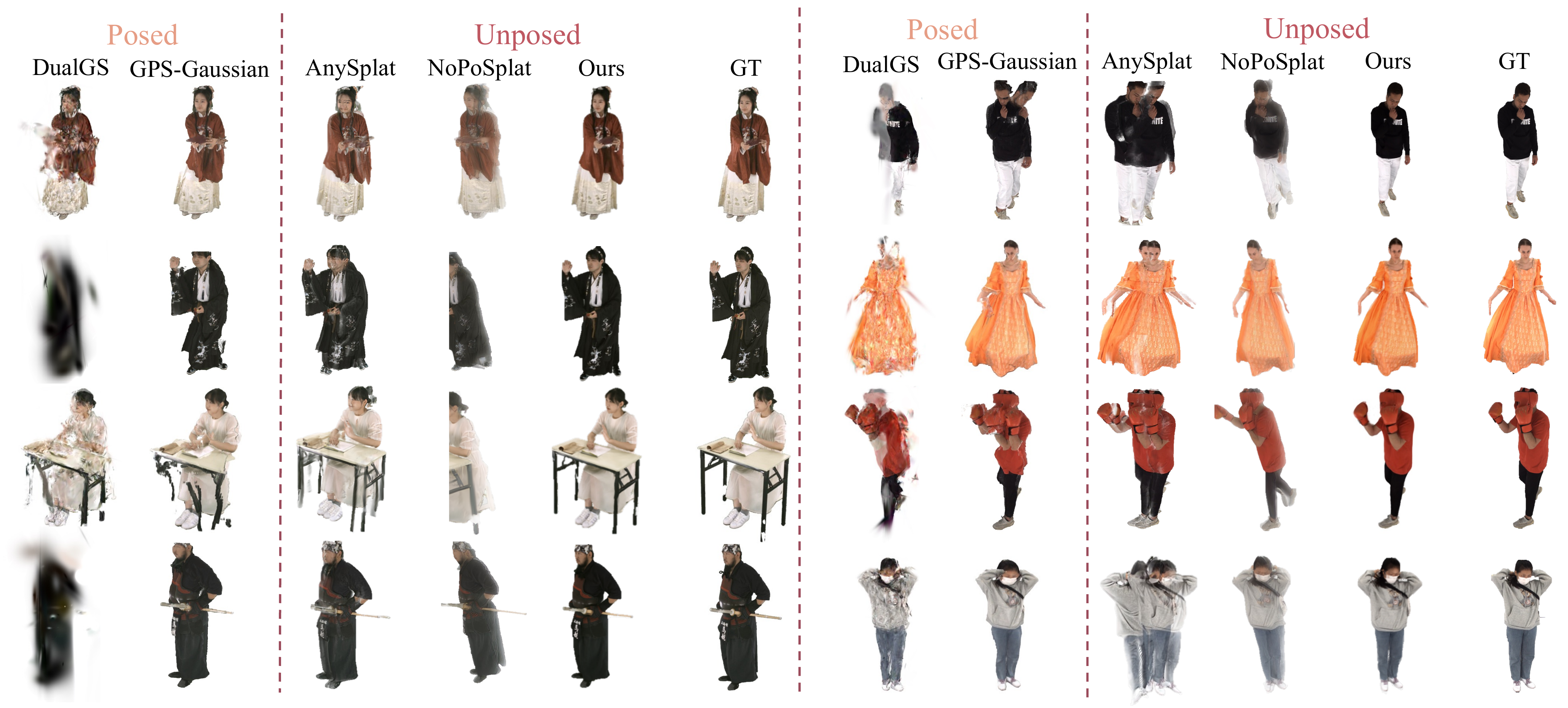}
    \caption{Qualitative results of 3D reconstruction on test sets. Our \textit{\methodname} exhibits more stable synthesized novel-view images against artifacts, including blur, ghosting, and shape distortion.}
    \label{fig:staticcomp}
\end{figure}

\textbf{Evaluation on 3D Reconstruction and Novel-View Synthesis.}
As demonstrated in Tab.~\ref{3Dexp}, our model outperforms all baseline methods across all metrics. Specifically, \textit{\methodname} surpasses previous pose-free feed-forward 3D reconstruction models, i.e., NoPoSplat and AnySplat, by up to +2.28 dB in PSNR. This significant performance gap stems not only from artifacts caused by multi-view mismatches and geometric inaccuracies but also from the inability of these methods to generate 3D models and camera configurations that align with the ground-truth scale and camera parameters. 
Moreover, the extensive white background in human images further complicates the synthesis of plausible renderings, even with the camera optimization procedure described in Supplementary C.
Compared to posed feed-forward 3D reconstruction models, \textit{\methodname} also exceeds the previous state-of-the-art method, GPS-Gaussian, by up to +2.90 dB in PSNR, which originates from \textit{\methodname}'s ability to produce more geometrically faithful reconstructions of challenging regions such as hands, heads, and accessories. Qualitative results in Fig.~\ref{fig:staticcomp} further confirm that \textit{\methodname} delivers more photorealistic novel views without artifacts such as blur, ghosting, or shape distortion.

\begin{figure}[t]
    \centering
    \includegraphics[width=1.\linewidth]{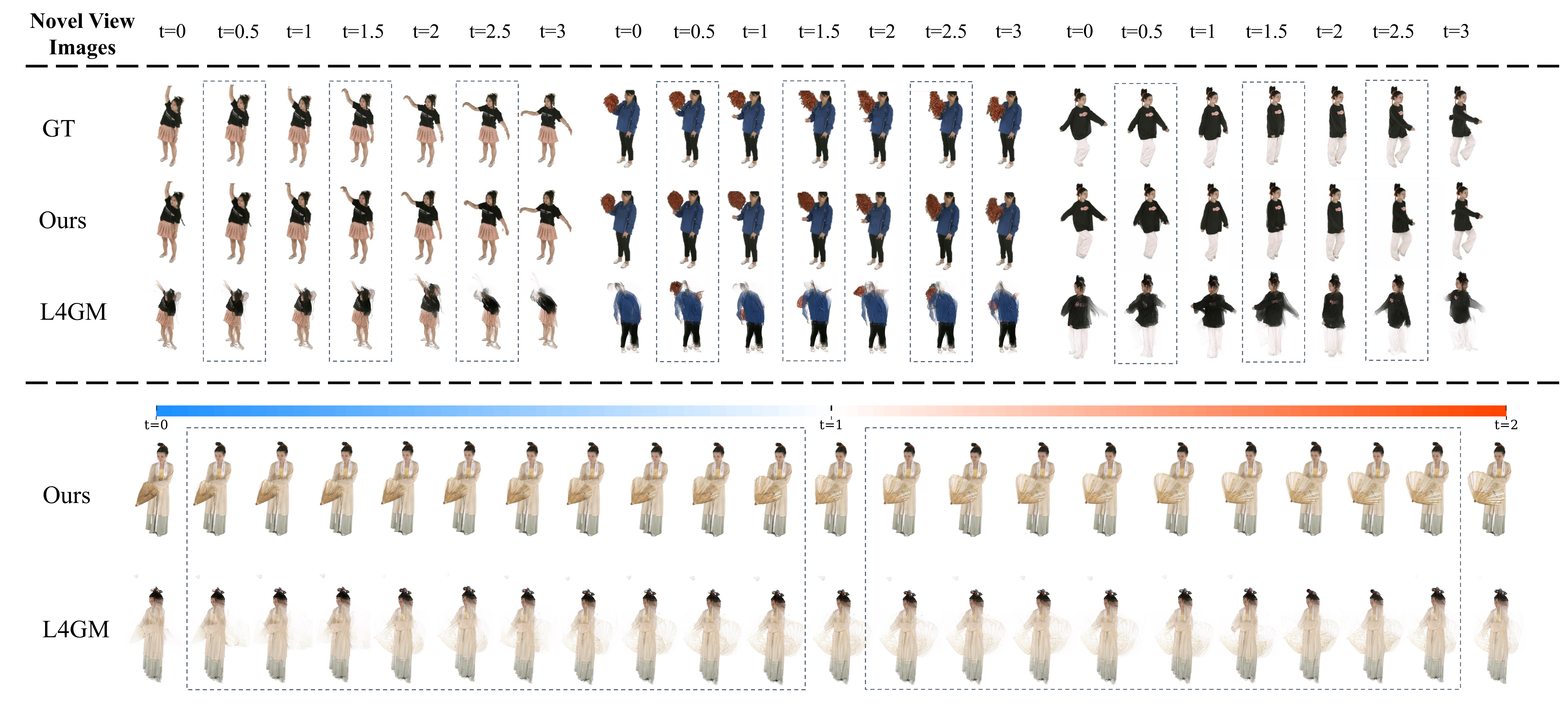}
    \caption{Qualitative results of 4D reconstruction on novel-view and novel-time synthesis. \textit{\methodname} accurately reconstructs 3DGS at input frames while generating plausible intermediate 3DGS \textit{at any time} with high-fidelity rendering quality (images in dashes).
    }
    \label{fig:dynamiccomp}
\end{figure}

\begin{figure}
    \centering
    \includegraphics[width=0.8\linewidth]{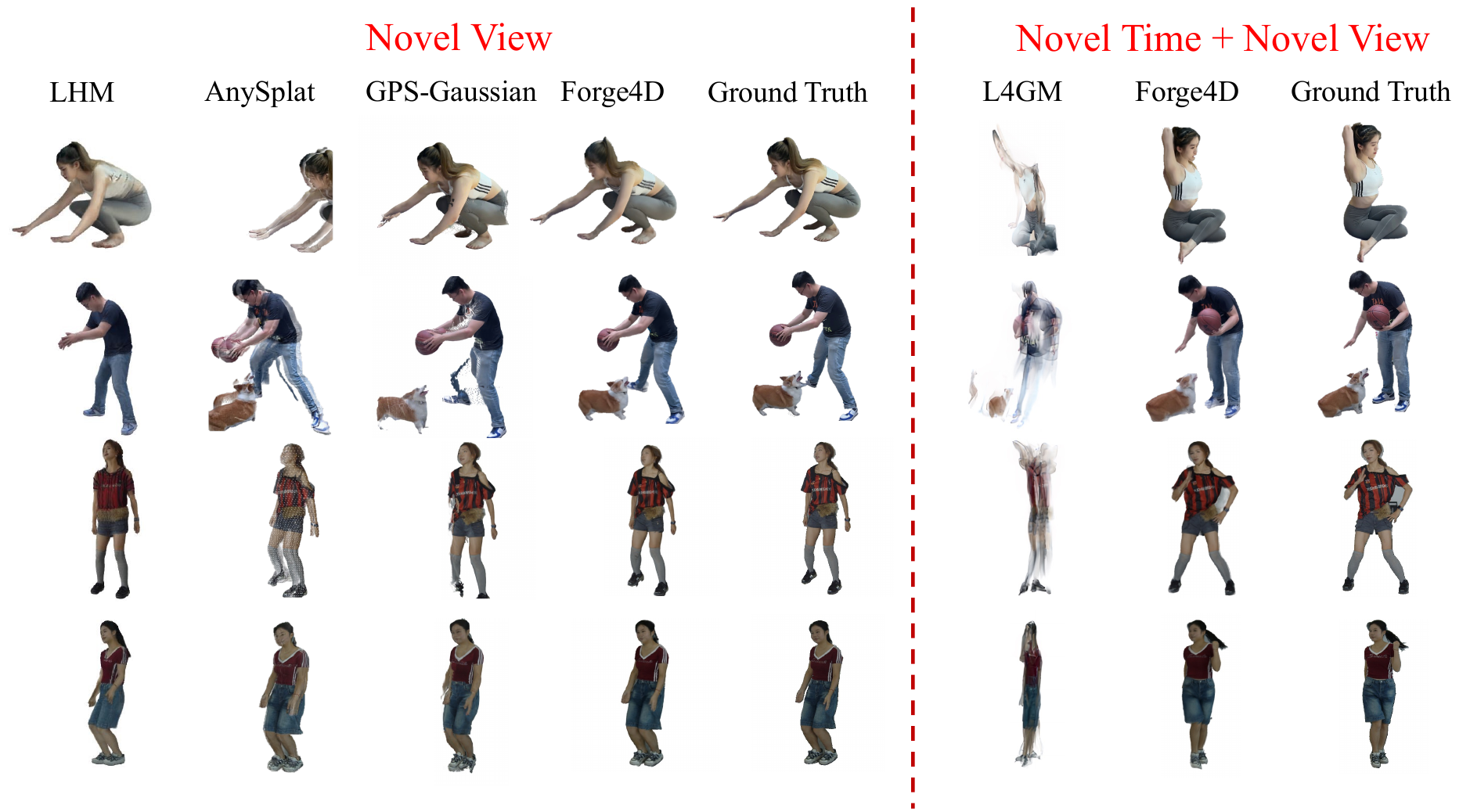}
    \caption{Qualitative results on in-the-wild datasets of the human-centric methods.}
    \label{fig:realworld}
\end{figure}
\noindent\textbf{Evaluation on 4D Reconstruction and Novel-Time Synthesis.}
Quantitative and qualitative results in Tab.~\ref{4Dexp} and Fig.~\ref{fig:dynamiccomp} demonstrate \textit{\methodname}'s effectiveness in novel-time synthesis. The model exhibits strong capabilities in both generating 3D Gaussian assets and interpolating 3DGS for arbitrary intermediate timestamps between key frames. Our approach outperforms the previous state-of-the-art feed-forward 4D reconstruction model L4GM on both datasets with performance gains of up to +12.57 dB in PSNR.
In comparison with optimization-based methods, our method surpasses all previous approaches, as these methods fail to generate reasonable novel views under such sparse-view conditions. The optimization-based techniques struggle with the limited input views, while our feed-forward approach maintains robust performance even with sparse camera arrangements.

\begin{table}[t]
    \centering
    \begin{minipage}{0.40\linewidth}
        \centering
        \caption{In-the-wild accessments.}
        \resizebox{\linewidth}{!}{
            \begin{tabular}{ccccccc}
                \hline
                \multirow{2}{*}{Method} & \multicolumn{3}{c}{\textbf{Novel-View}} & \multicolumn{3}{c}{\textbf{Novel-Time}} \\
                \cline{2-7}
                & PSNR$\uparrow$ & SSIM$\uparrow$ & LPIPS$\downarrow$ & PSNR$\uparrow$ & SSIM$\uparrow$ & LPIPS$\downarrow$ \\ \hline
                LHM          & 13.3687 & 0.7867 & 0.3942 & - & - & - \\
                GPS-Gaussian & 25.6316 & 0.9065 & 0.1019 & - & - & - \\
                AnySplat     & 23.9645 & 0.8930 & 0.1502 & - & - & - \\
                L4GM         & 15.4771 & 0.8494 & 0.2111 & 14.8831 & 0.8170 & 0.1951 \\
                Forge4D      & 26.2735 & 0.9160 & 0.0917 & 25.4726 & 0.9009 & 0.1025 \\ \hline
            \end{tabular}
        }
        \label{tab:realworld}
    \end{minipage}
    \hfill
    \begin{minipage}{0.2\linewidth}
        \centering
        \caption{Metric Scale prediction evaluation.}
        \resizebox{\linewidth}{!}{
            \begin{tabular}{cc}
                \toprule
                \textbf{Method} & \textbf{Point Distance} \\ \hline
                MoGe-2 & 0.3309m \\
                Ours   & \textbf{0.0264m} \\ \bottomrule
            \end{tabular}
        }
        \label{exp:metric}
    \end{minipage}
    \hfill
    \begin{minipage}{0.35\linewidth}
        \centering
        \caption{Motion prediction evaluation.}
        \resizebox{\linewidth}{!}{
            \begin{tabular}{ccc}
                \toprule
                \textbf{Method} & \textbf{Motion Error} & \textbf{Point Distance} \\ \hline
                POMATO & 0.01274 & 0.7555 \\
                Ours   & \textbf{0.00953} & \textbf{0.0215} \\ \bottomrule
            \end{tabular}
        }
        \label{exp:motion}
    \end{minipage}
\end{table}
\noindent\textbf{Evaluation on In-the-wild Dataset.} We evaluate the in-the-wild applicability of \textit{\methodname} on two casually captured datasets: SelfCap~\cite{freetimegs} and VVC~\cite{vvc}. From SelfCap, we select yoga and corgi sequences that effectively capture full-body human motion with objects, while using the complete validation and test sets from VVC. This comprehensive evaluation covers 9 diverse human-object interaction scenarios, representing a wide spectrum of real-world situations. We select 4 cameras with diverse viewing angles ranging from 20 to 70 degrees as model inputs, corresponding to their spatial arrangement in each test sequence.
We provide quantitative comparisons with GPS-Gaussian~\cite{gpsgaussian} in Tab.~\ref{tab:realworld} and qualitative comparisons in Fig.~\ref{fig:realworld}.

\noindent\textbf{Evaluation on Dense Motion Prediction}. 
We evaluate our motion prediction module using our MetaHuman4D dataset containing multi-view images and dense ground-truth motions derived from mesh correspondences for each frame. The predicted 3D motions are compared against the GT 3D motions using L2 distance, and benchmarked against the state-of-the-art dense motion prediction model POMATO~\cite{pomato}. Additionally, we report the retargeted point distance to the GT target-time mesh, where \textit{\methodname} consistently generates plausible outputs while POMATO fails to produce reasonable human geometry. Quantitative results in Tab.~\ref{exp:motion} demonstrate the effectiveness of our motion prediction framework. Further details are provided in Supplementary B.

\noindent\textbf{Evaluation on Metric Scale Prediction}. \textit{\methodname} is able to recover real-world metric scale points as a byproduct of the scale prediction header supervised with \textit{metric gauge}. We evaluate the performance of \textit{\methodname} in metric scale recovering by measuring the mean distance of the predicted points to the GT human mesh on MetaHuman4D, which results in a $0.02$ m error on average.
A comparison with MoGe-2 is made in the same metric, with the results presented in Tab.~\ref{exp:metric}. \textit{\methodname} outperforms MoGe-2 in mean distance to ground-truth mesh, primarily due to MoGe-2's inability to effectively align multi-view point correspondences.

\subsection{Ablation Study}
\label{ablation}

\begin{figure}[t]
    \centering
    \includegraphics[width=0.9\linewidth]{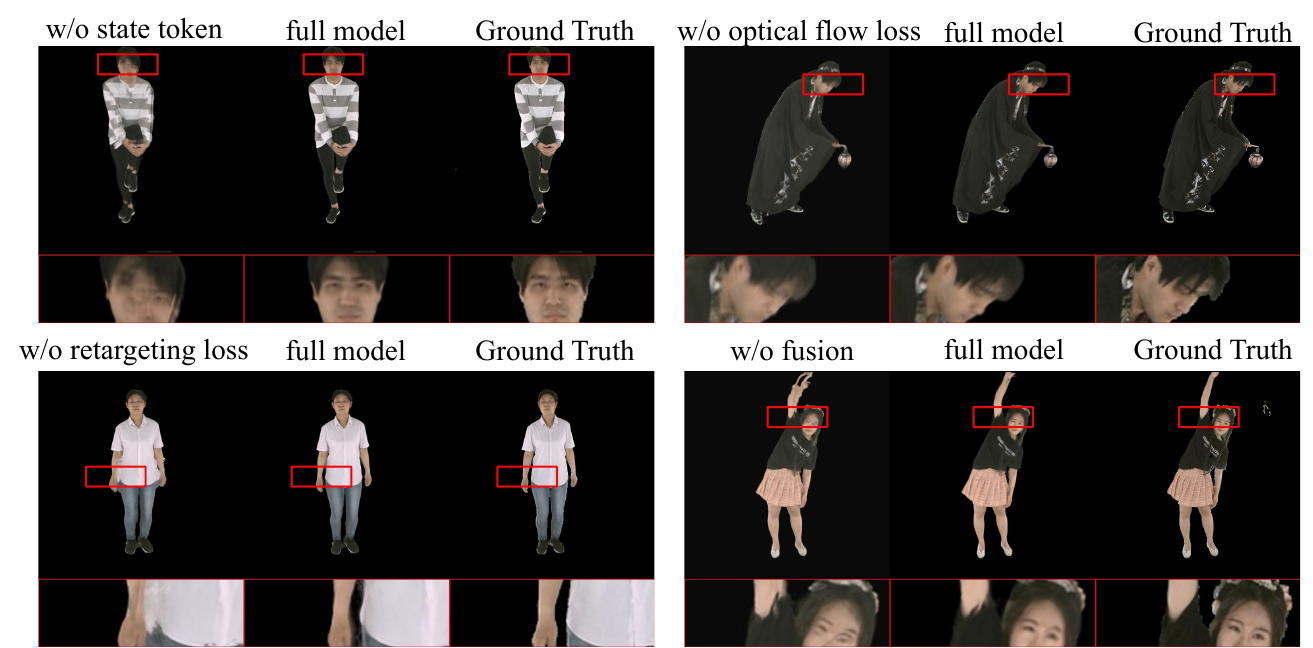}
    \caption{Visualizations on the ablation study.}
    \label{fig:ablation}
\end{figure}

\noindent\textbf{Ablations on Methodology}. We show the effectiveness of the proposed \textit{metric gauge}, the \textit{retargeting loss}, the \textit{occlusion-aware optical flow loss}, and the Gaussian fusion process in Tab.~\ref{abla:components}. A significant performance gap is observed when the \textit{retargeting loss} and the occlusion-aware \textit{optical flow loss} are replaced with direct supervision on the novel timestamps, and the training process will lead to a collapse when the \textit{metric gauge} alignment is missing. Additionally, directly averaging the 3DGS from the nearby 2 frames will also lead to a suboptimal novel-view image quality. We provide novel-view and novel-time visualizations for the ablation study in Sec. 4.3 in Figure~\ref{fig:ablation}. The results demonstrate that without state token updates, temporal misalignment occurs at novel timestamps, while the absence of either retargeting loss or optical flow loss leads to unreliable human motion estimation, and vanilla averaging the Gaussian attributes leads to temporal redunancy and flickering issues. We also include a comparison of the Temporal Luminance Standard Deviation (TL-STD$\downarrow$) to illustrate these flickering issues, where a higher TL-STD indicates more noticeable brightness jitters. Without Gaussian Fusion, the average TL-STD (0.9446) is 75.6\% higher than that with Gaussian Fusion (0.5379).

\begin{table}[t]
    \centering
    \begin{minipage}{0.45\linewidth}
        \centering
        \caption{Ablation on model speed.}
        \scriptsize
        \resizebox{0.99\linewidth}{!}{
            \begin{tabular}{lcc}
                \toprule
                Module  & Delay & Frame Rate \\ \hline
                Key-frame Reconstruction         & 176.50 ms    & - \\
                Motion Prediction         & 47.77 ms    & - \\
                Interpolation (10 Steps)         & 1.46 ms    & - \\ \hline
                Full Model         & 224.27 ms    & 4.45 FPS  \\
                \textit{+Interpolate 10 Steps}       & 225.10 ms &  44.42 FPS \\
                \textit{+Interpolate 20 Steps}       & 226.01 ms &  88.48 FPS \\ \bottomrule
            \end{tabular}
        }
        \label{abla:time}
    \end{minipage}
    \hfill
    \begin{minipage}{0.52\linewidth}
        \centering
        \caption{Ablation on different components.}
        \scriptsize
        \resizebox{\linewidth}{!}{
            \begin{tabular}{llccc}
                \toprule
                \textbf{Evaluation Mode}  & \textbf{Variants} & \textbf{PSNR}$\uparrow$ & \textbf{SSIM}$\uparrow$ & \textbf{LPIPS}$\downarrow$ \\
                \midrule
                \multirow{2}{*}{Static Novel View} & \textbf{Full Model} & 29.8167 & 0.9606 & 0.0542 \\
                  & w/o \textit{gauge aligning} & 13.2884 & 0.1194 & 0.2184 \\
                \midrule
                \multirow{5}{*}{\makecell{Dynamic \\ Novel Time \\ + Novel View}} & \textbf{Full Model} & 29.0378 & 0.9566 & 0.0535 \\
                  & w/o \textit{state token} & 28.5555 & 0.9513 & 0.0592 \\
                  & w/o \textit{retargeting loss} & 28.4124 & 0.9530 & 0.0573 \\
                  & w/o \textit{optical flow loss} & 28.8676 & 0.9551 & 0.0563 \\
                  & w/o \textit{Gaussian Fusion} & 27.3459 & 0.9346 & 0.0683 \\
                \bottomrule
            \end{tabular}
        }
        \label{abla:components}
    \end{minipage}
\end{table}

\noindent\textbf{Ablations on Model Speed}. We evaluate the inference speed of \textit{\methodname} on an NVIDIA H200 Tensor Core GPU, with the results reported in Tab.~\ref{abla:time}. The key-frame reconstruction requires 176.50 ms per inference, the motion prediction module takes 47.77 ms, and the intermediate-time interpolation for 10 frames adds 1.46 ms, resulting in a total latency of 224.27 ms per input frame pair. Given our model's capability for arbitrary-length interpolation, the effective output frame rate reaches 44 FPS when generating 10 interpolated frames per input interval.

%% file: sec/5_conclusion.tex
\section{Conclusion}
We propose \textit{\methodname}, the first feed-forward model for 4D human-object reconstruction from uncalibrated sparse-view videos. Our approach simplifies the problem by decomposing it into streaming real-world metric-scale 3D Gaussian reconstruction and dense motion prediction for novel-time synthesis. \textit{\methodname} achieves state-of-the-art novel-view synthesis quality and enables interpolation to arbitrary timestamps while maintaining plausible intermediate representations. Nevertheless, \textit{\methodname} exhibits certain limitations. The performance degrades in the presence of large motions or longer inter-frame intervals, primarily due to reduced inter-frame correspondences and violations of the consistent motion assumption. We identify these limitations as directions for future work, focusing on improving motion modeling under extreme displacements and optimizing computational efficiency for better interactivity.

%% file: sec/x_appendix.tex
\clearpage
\setcounter{page}{1}
\title{Supplementary Material for \logo \textcolor{color1}{For}\textcolor{color2}{ge}\textcolor{color3}{4D}: Feed-\textcolor{color1}{For}ward \textcolor{color3}{4D} Human Reconstruction and Interpolation from Uncalibrated Sparse-View Videos}

\titlerunning{Forge4D}

\author{Yingdong Hu$^{\ast}$\inst{1}\orcidlink{0009-0002-8708-1001}
\and
Yisheng He$^{\ast}$\textsuperscript{\Letter}\inst{2}\orcidlink{0000-0002-4029-5494} \and
Jinnan Chen\inst{3}\orcidlink{0009-0007-9964-2132} \and
Weihao Yuan\inst{2}\orcidlink{0000-0002-1362-3747} \and
Kejie Qiu\inst{2}\orcidlink{0000-0002-2516-3020}  \and
Zehong Lin\inst{4}\orcidlink{0000-0002-9503-2464}  \and
Siyu Zhu\inst{5}\orcidlink{0000-0003-0293-0044}  \and
Zilong Dong\inst{2}\orcidlink{ 0000-0002-6833-9102}  \and
Steven Hoi\inst{2}\orcidlink{0000-0002-4584-3453}\and
Jun Zhang\inst{1}\orcidlink{0000-0002-5222-1898}}

\authorrunning{Y.~Hu et al.}

\institute{The Hong Kong University of Science and Technology, Hong Kong SAR, China\and
Tongyi Lab, Alibaba Group, Hangzhou, China
\and
National University of Singapore, Singapore
\and
School of Data Science, Lingnan University, Hong Kong SAR, China\and
Fudan University, Shanghai, China\\
\email{yingdong.hu@connect.ust.hk, ethanheysh@gmail.com}
}

\maketitle
\setcounter{figure}{7}
\setcounter{table}{7}
\footnotetext[1]{$^{\ast}$ indicates equal contribution.}
\footnotetext[2]{\textsuperscript{\Letter} Corresponding author. Contact:\texttt{ethanheysh@gmail.com}}

\section{Overview}
In Sec.~\ref{appendix:motion}, we detail the synthetic MetaHuman4D dataset, including its composition and our ground-truth motion annotation methodology.  Sec.~\ref{appendix:implementation} provides additional training and evaluation specifics, along with visualizations of \textit{\methodname}'s motion prediction results. Sec.~\ref{appendix:moreablations} presents further ablations concerning video duration, temporal intervals, and camera count, and more visualization on the ablation study in Sec.~\ref{appendix:results} provides more experimental results on in-the-wild assessments, comparisons with template-based methods, speed test. Sec.~\ref{appendix:metric} illustrates \textit{\methodname}'s metric scale prediction capabilities. 
Sec.~\ref{appendix:ethic_state} discusses potential ethical concerns.

\section{Our Synthesis MetaHuman4D Dataset}
\label{appendix:motion}
Since there are no ground truth annotations of dense motion and metric scale in the captured real-world dataset, we construct a synthesis dataset with ground truth annotations for the evaluation of different methods on these two tasks.

\begin{figure*}
    \centering
    \includegraphics[width=\linewidth]{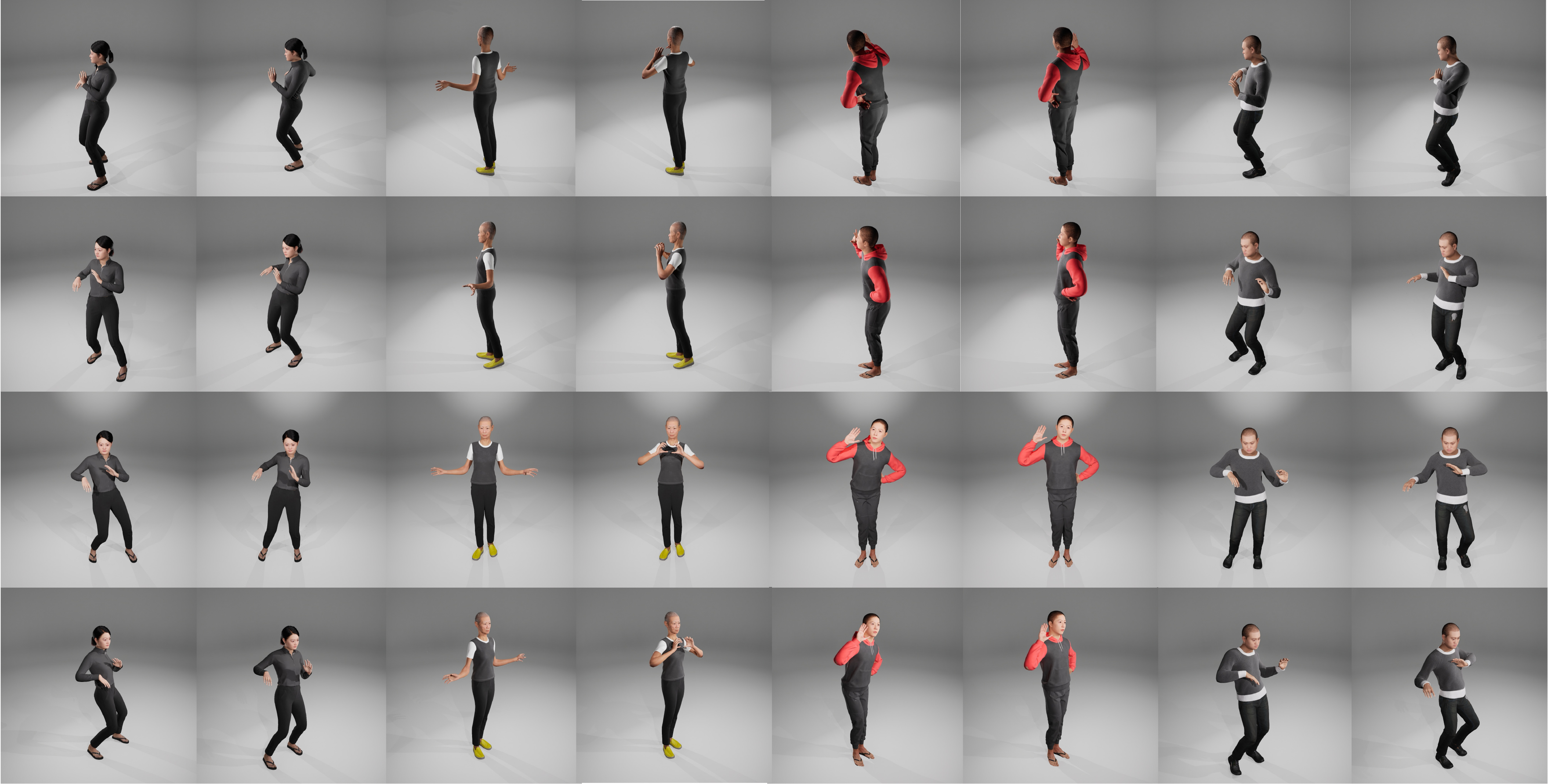}
    \caption{Sampled examples of our MetaHuman4D test set.}
    \label{fig:testset}
\end{figure*}

\noindent\textbf{Dataset details.}
Our synthesized test set contains 11 different identities and 7 motion types. For each person, we select a motion sequence to animate it and render dynamic videos from 48 views with the Unreal rendering engine and save the GT mesh model of the human at each given timestamp. The diversity of this dataset is at the same scale as the commonly used test set from DNA-rendering, which covers 10 identities. We visualize some samples in Fig.~\ref{fig:testset}.

\noindent\textbf{Ground truth motion annotation.} We derive ground-truth motion from sequential human meshes by computing the displacement of corresponding points between consecutive timestamps. Specifically, for a point $\bm{\bar{x}}^t_i$ on the mesh at time $t$, we obtain its backward motion as $\bar{\bm{m}}^t_1 = \bm{\bar{x}}^{t-1}_i - \bm{\bar{x}}^t_i$ and its forward motion as $\bar{\bm{m}}^t_2 = \bm{\bar{x}}^{t+1}_i - \bm{\bar{x}}^t_i$. During evaluation, we establish correspondences between predicted points and ground-truth mesh points, then compute the motion error between the predicted motions ${\bm{m}}^t_{1,2}$ and the ground-truth motions ${\bar{\bm{m}}}^t_{1,2}$ for each matched point pair.

\section{Implementation Details}
\label{appendix:implementation}
\noindent\textbf{Training detail of the three training stages.} Stage 1 employs a learning rate initialized at $5\times10^{-5}$ for the Gaussian and scale heads, and $1\times10^{-5}$ for other components. Stage 2 initializes with Stage 1 weights, except for the randomly initialized state attention module and state tokens. During this stage, the state attention module, state tokens, scale head, Gaussian head, and pose head are optimized with a $5\times10^{-5}$ initial learning rate, while other components remain frozen. Stage 3 initializes with Stage 2 weights, with randomly initialized motion blocks, Gaussian fusion, and a zero-initialized motion head module, all trained at $5\times10^{-5}$ initial learning rate, while all other parameters remain frozen. All stages use a consistent batch size of 8 and are trained on 8 H20 GPUs for 100,000 iterations each, and the learning rate is linearly decreasing to $1\times10^{-5}$ at the end of the training stage. Hyperparameters are empirically set as: $\lambda_\text{S} = 0.25$, $\lambda_\text{L} = 0.25$, $\tau = 0.05$, $\bm{r}_i^t = 0.1\cdot||\bm{\mu}_i^{t\to t-1}||_{2,2}+0.5$. $||\cdot||_{2,2}$ represents the L2 norm on the second dimension of the tensor.

\noindent\textbf{Finetuning Camera Parameter Estimation.} The metric gauge's accuracy depends on the camera head output adhering to the scaling relationship defined in Sec.~3.2, as camera position errors can compromise the reliability of calculated metric gauge during training. To address this, we finetune the camera head using both photometric and metric gauge losses, ensuring predicted camera parameters maintain consistent scaling throughout training stage.

\noindent\textbf{Metric scale recovery during inference.} During inference, metric-scale 3D points are recovered by dividing the output 3D points by the predicted metric gauge $\hat{p}_{\text{gauge}}$, with the corresponding adjustments applied to the scale and motion attributes of 3DGS for novel-view and novel-time rendering.

\noindent\textbf{Evaluation of pose-free methods.} Same with previous works~\cite{noposplat,instantsplat,nerf_min}, we optimize novel-view camera positions while keeping the 3DGS fixed for pose-free methods (NoPosplat~\cite{noposplat}, AnySplat~\cite{anysplat}, \textit{\methodname}) to address the inherent ambiguity that multiple 3D configurations can explain the same 2D observations. Importantly, this optimization is solely performed for evaluation and is not required during actual deployment for novel-view and novel-time rendering.

\noindent\textbf{Evaluation details of different baselines on novel-view synthesis of static reconstruction.} In Sec.~4.2, all 3D methods are evaluated using frame-wise reconstruction and rendering, while DualGS and Queen are optimized using full video sequences. We disable state tokens and state attention blocks in \methodname~for fair comparisons. Note that DualGS cannot synthesize novel-time 3D scenes.

\noindent\textbf{Evaluation details of different baselines on novel-time and novel-view synthesis.} All 4D methods are evaluated at a sampling rate of 2, which is holding out 1 timestamp between every 2 input timestamps for novel-time and novel-view evaluation. Since L4GM can only take a maximum input length of 8 timestamps due to the GPU memory limitation, we compare it with our model at the same input video length of 8 timestamps for fair comparison. However, we claim that our model can be extended to arbitrary length of videos without suffering from memory accumulation issues, thanks to the effectiveness of the state token embeddings.

\begin{figure*}
    \centering
    \includegraphics[width=\linewidth]{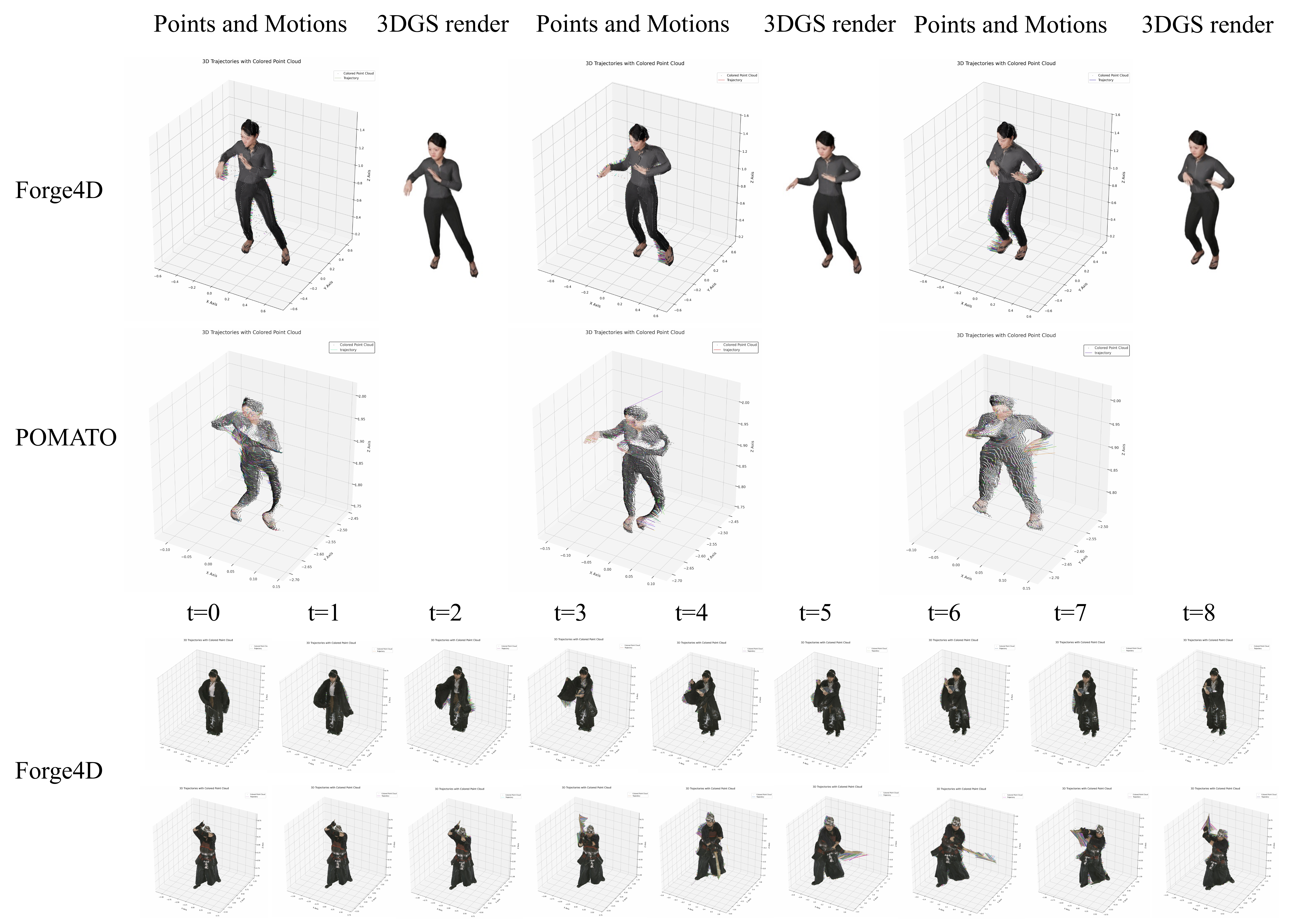}
    \caption{Dense motion prediction visualization. Zoom in to see more details.}
    \label{fig:motion}
\end{figure*}

\noindent\textbf{Evaluation details of dense motion prediction.}
For our experimental setup, we select 4 sparse views from the rendered videos as input to each compared method for human-object reconstruction. To ensure a fair quantitative evaluation, we first align the predicted 3D points at the initial timestamp with the ground-truth mesh from the MetaHuman4D dataset using scaling, translation, and rotation transformations. This alignment step eliminates errors arising from scale and coordinate system discrepancies, especially for POMATO~\cite{pomato}. Subsequently, for each predicted 3D point, we identify the closest point on the ground-truth mesh, establishing a correspondence for motion evaluation. The motion accuracy is quantified by computing the L2 distance between the motion vectors of each predicted point and its matched ground-truth point. Additionally, we report the mean distance between the deformed predicted 3D points and their corresponding ground-truth points at the target timestamp. We visualize the predicted 3D points and sampled motions in Fig.~\ref{fig:motion}.

\noindent\textbf{Evaluation details of metric measurement.}
We evaluate \textit{\methodname} and MoGe-2~\cite{moge2} using identical videos with dense motion annotations. For each predicted 3D point, we compute its minimum distance to the ground-truth mesh surface as the per-point prediction error. The mean distance is calculated directly without additional alignment procedures, beyond the necessary transformation of predicted points from camera coordinates to world coordinates. For MoGe-2, we independently calculate the prediction error for each view, and make an average on the 4 input views.

\section{More Ablations}
\label{appendix:moreablations}

\noindent\textbf{Ablations on Video Duration.} We evaluate our model under different video durations and report the metrics of novel-view and novel-time image quality, in 
terms of PSNR, SSIM, and LPIPS. The results in Tab.~\ref{abla:timestamps} show that the duration of the video has a limited impact on the model performance.

\begin{table}[htbp]
    \centering
    \begin{minipage}{0.32\linewidth}
        \centering
        \caption{Ablation on camera settings.}
        \scriptsize
        \resizebox{\linewidth}{!}{
            \begin{tabular}{c|ccc}
                \hline
                Cam. Num. & 2 & 4 & 5 \\ \hline
                PSNR$\uparrow$   & 27.1316 & 29.0378 & 28.1337 \\
                SSIM$\uparrow$   & 0.9384  & 0.9566  & 0.9471  \\
                LPIPS$\downarrow$ & 0.0648  & 0.0535  & 0.0613  \\ \hline
            \end{tabular}
        }
        \label{abla:camera}
    \end{minipage}
    \hfill
    \begin{minipage}{0.32\linewidth}
        \centering
        \caption{Ablation on the sampling rate.}
        \scriptsize
        \resizebox{\linewidth}{!}{
            \begin{tabular}{c|ccc}
                \hline
                Sampling Rate & 2 & 4 & 8 \\ \hline
                PSNR$\uparrow$   & 29.0378 & 27.9129 & 25.9673 \\
                SSIM$\uparrow$   & 0.9566  & 0.9538  & 0.9344  \\
                LPIPS$\downarrow$ & 0.0535  & 0.0604  & 0.0767  \\ \hline
            \end{tabular}
        }
        \label{abla:sampling}
    \end{minipage}
    \hfill
    \begin{minipage}{0.32\linewidth}
        \centering
        \caption{Ablation on the video length.}
        \scriptsize
        \resizebox{\linewidth}{!}{
            \begin{tabular}{c|ccc}
                \hline
                Timestamps & 8 & 16 & 32 \\ \hline
                PSNR$\uparrow$   & 29.0378 & 29.3932 & 29.3615 \\
                SSIM$\uparrow$   & 0.9566  & 0.9591  & 0.9591  \\
                LPIPS$\downarrow$ & 0.0535  & 0.0519  & 0.0519  \\ \hline
            \end{tabular}
        }
        \label{abla:timestamps}
    \end{minipage}
\end{table}
\noindent\textbf{Ablations on Time interval and Interpolation rate.} \textit{\methodname} is trained by sampling a timestamp every two timestamps as novel-time supervision frames, which we refer to as a sampling rate of two. We show that this supervision strategy is sufficient and that \textit{\methodname} can generalize to longer time intervals. In addition, more plausible intermediate frames can be acquired by adjusting the interpolation rate. We carried out metric calculations for larger sampling rates of 4 and 8. The quantitative result in Tab.\ref{abla:sampling} indicates that the 
rendered novel-view quality is preserved even when a longer duration is held between the two input timestamps. There is only a reasonable drop in PSNR of at most 3.04 dB, which is mainly because the linear velocity assumption is hard to maintain under a longer duration.

\noindent\textbf{Ablations on Camera number.} Although \textit{\methodname} is trained with a consistent configuration of 4 input views, the model demonstrates strong generalization capability to arbitrary numbers of input cameras. As shown in Table~\ref{abla:camera}, which evaluates novel-view and novel-time synthesis quality with 2, 4, and 5 input views, the performance degradation remains minimal, with a maximum decrease of only 1.90 dB in PSNR. This robustness stems from the inherent stability of our backbone architecture and the fact that cross-frame information aggregation is largely decoupled from the final 3D Gaussian prediction heads, allowing the model to adapt effectively to varying numbers of input views.

\section{More Results}
\label{appendix:results}
\noindent\textbf{Speed Comparison.} We compare reconstruction latency and frame rates with existing methods in Tab.~\ref{abla:speed}. The primary speed limitation for template-based method LHM stems from its SMPL-X parameter estimation module, Multi-HMR~\cite{multi-hmr}. All measurements are obtained on an H20 GPU with 4 input images at 518×518 (512×512 for methods specified in Sec. 4.1) resolution, except LHM+Multi-HMR, which only takes monocular input and is non-trivial to adapt to multi-view setting. The delay time measures the duration from first frame input to first frame output, while the frame rate indicates the output frames per second during subsequent inference.

\begin{table}[h]
    \centering
    \scriptsize
    \caption{Speed\&Quality comparison.}
    \resizebox{0.7\linewidth}{!}{\begin{tabular}{cccccc}
        \hline
        Method &w. Cam. pose&Type& Delay & Frame Rate & PSNR \\ \hline
        GPS-Gaussian&Yes      &3D&0.0703s&14.22FPS&24.2963\\ 
        Queen &Yes &3D&30.14s&0.725FPS&15.4966\\
        LHM+Multi-HMR   &\noindent\textbf{No}   &3D&5.330s&22.22FPS&13.3687\\ 
        AnySplat &\noindent\textbf{No} &3D &0.4636s&2.157FPS&26.1157\\ \hline
        D-3DGS&Yes &\noindent\textbf{4D} &8min38s&0.1453FPS&20.9158\\
        SpaceTimeGS&Yes &\noindent\textbf{4D} &1h52min&0.022FPS&17.2189\\
        L4GM&Yes &\noindent\textbf{4D} &0.2231s&71.72FPS&18.0325\\
        Forge4D (+20 interp.) &\noindent\textbf{No}     &\noindent\textbf{4D}& 0.5035s&39.72FPS&\noindent\textbf{29.0378} \\ \hline
    \end{tabular}}
    \label{abla:speed}
\end{table}

\section{Metric Measurement}
\label{appendix:metric}
\begin{figure*}
    \centering
    \includegraphics[width=1.\linewidth]{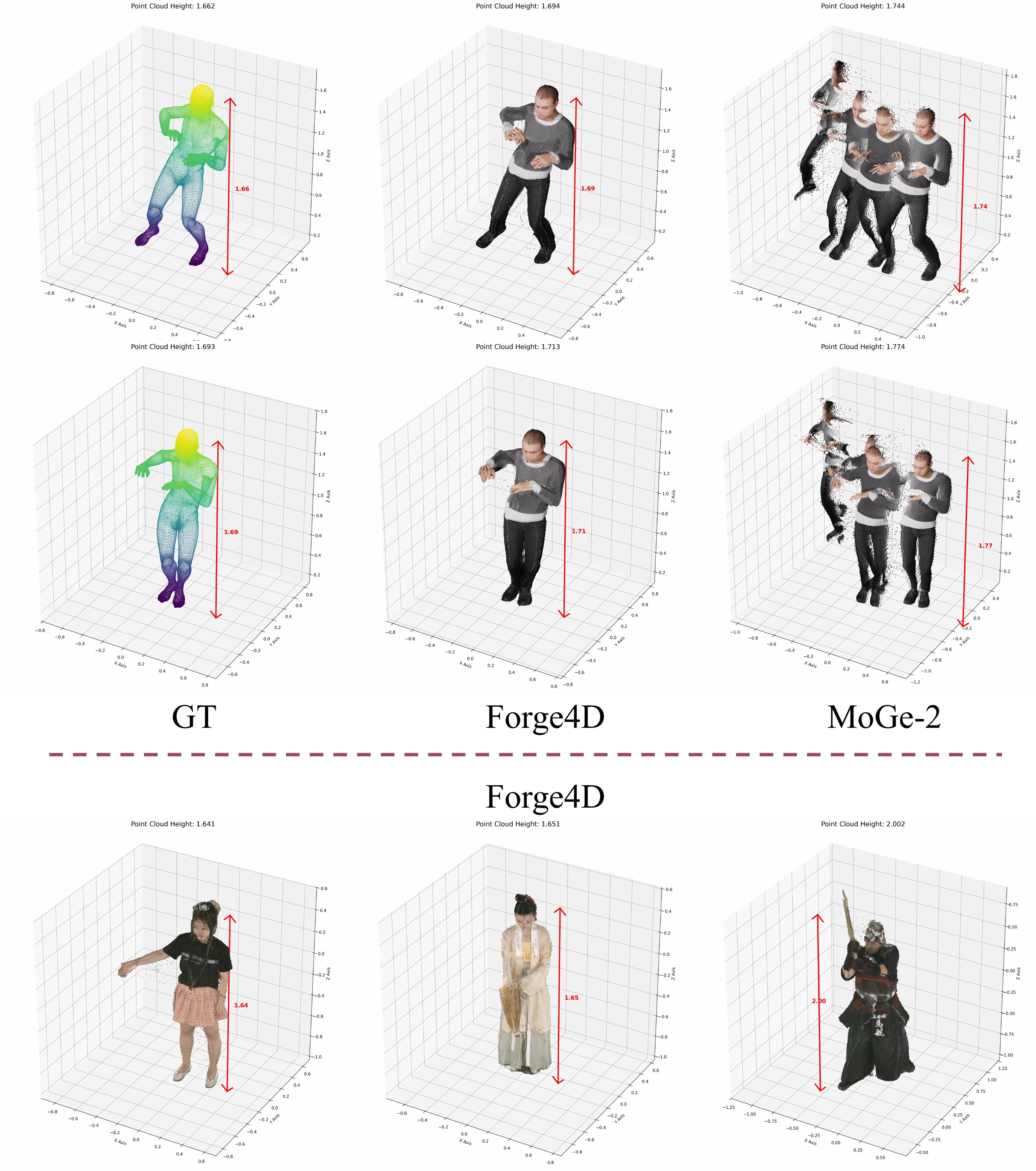}
    \caption{Samples of points prediction in metric scale.}
    \label{fig:metric}
\end{figure*}

We provide additional human body metric measurement results in Figure~\ref{fig:metric}, demonstrating that \textit{\methodname} achieves accurate metric-scale reconstruction across both synthetic datasets and real-world captures, validating its robustness in practical measurement applications.

\section{Ethics Statement.} 
\label{appendix:ethic_state}
The research utilizes DNA-Rendering, Genebody, and the synthesized MetaHuman4D datasets, all employing properly consented data or synthetic human models to avoid privacy concerns. Although developed for beneficial applications, we acknowledge the potential misuse of this technology for creating synthetic media without consent and encourage the development of corresponding ethical guidelines and detection mechanisms. We believe the primary impact of this research will be positive, enabling new forms of human communication and content creation, while emphasizing the need for responsible development and deployment.